\documentclass[10pt,twocolumn,letterpaper]{article}
\usepackage[accsupp]{axessibility}
\usepackage{iccv}
\usepackage{times}
\usepackage{epsfig}
\usepackage{graphicx}
\usepackage{amsmath}
\usepackage{amssymb}
\usepackage{booktabs}
\usepackage{makecell}
\usepackage{float}
\usepackage{color}
\usepackage{dsfont}
\usepackage{mathrsfs}
\usepackage{bbm}
\usepackage{ulem}
\usepackage{bm}

\usepackage[pagebackref=true,breaklinks=true,letterpaper=true,colorlinks,bookmarks=false]{hyperref}

\iccvfinalcopy 

\def\iccvPaperID{****}

\ificcvfinal\fi

\begin{document}

\title{RecRecNet: Rectangling Rectified Wide-Angle Images by Thin-Plate Spline Model and DoF-based Curriculum Learning}

\author{Kang Liao$^{\dagger}$ \ \ Lang Nie$^{\dagger}$ \ \ Chunyu Lin\thanks{Corresponding author. $^{\dagger}$The first two authors contribute equally.} \ \ Zishuo Zheng \ \  Yao Zhao\\
Institute of Information Science, Beijing Jiaotong University\\
Beijing Key Laboratory of Advanced Information Science and Network\\
\tt\small {\{kang\_liao, nielang, cylin, yzhao\}@bjtu.edu.cn}}
\maketitle
\ificcvfinal\thispagestyle{empty}\fi

\begin{abstract}

The wide-angle lens shows appealing applications in VR technologies, but it introduces severe radial distortion into its captured image. To recover the realistic scene, previous works devote to rectifying the content of the wide-angle image. However, such a rectification solution inevitably distorts the image boundary, which changes related geometric distributions and misleads the current vision perception models. In this work, we explore constructing a win-win representation on both content and boundary by contributing a new learning model, \textit{i.e.}, Rectangling Rectification Network (RecRecNet). In particular, we propose a thin-plate spline (TPS) module to formulate the non-linear and non-rigid transformation for rectangling images. By learning the control points on the rectified image, our model can flexibly warp the source structure to the target domain and achieves an end-to-end unsupervised deformation. To relieve the complexity of structure approximation, we then inspire our RecRecNet to learn the gradual deformation rules with a DoF (Degree of Freedom)-based curriculum learning. By increasing the DoF in each curriculum stage, namely, from similarity transformation (4-DoF) to homography transformation (8-DoF), the network is capable of investigating more detailed deformations, offering fast convergence on the final rectangling task. Experiments show the superiority of our solution over the compared methods on both quantitative and qualitative evaluations. The code and dataset are available at \url{https://github.com/KangLiao929/RecRecNet}.

\end{abstract}

\begin{figure}[t]
\centering
\includegraphics[width=1\linewidth]{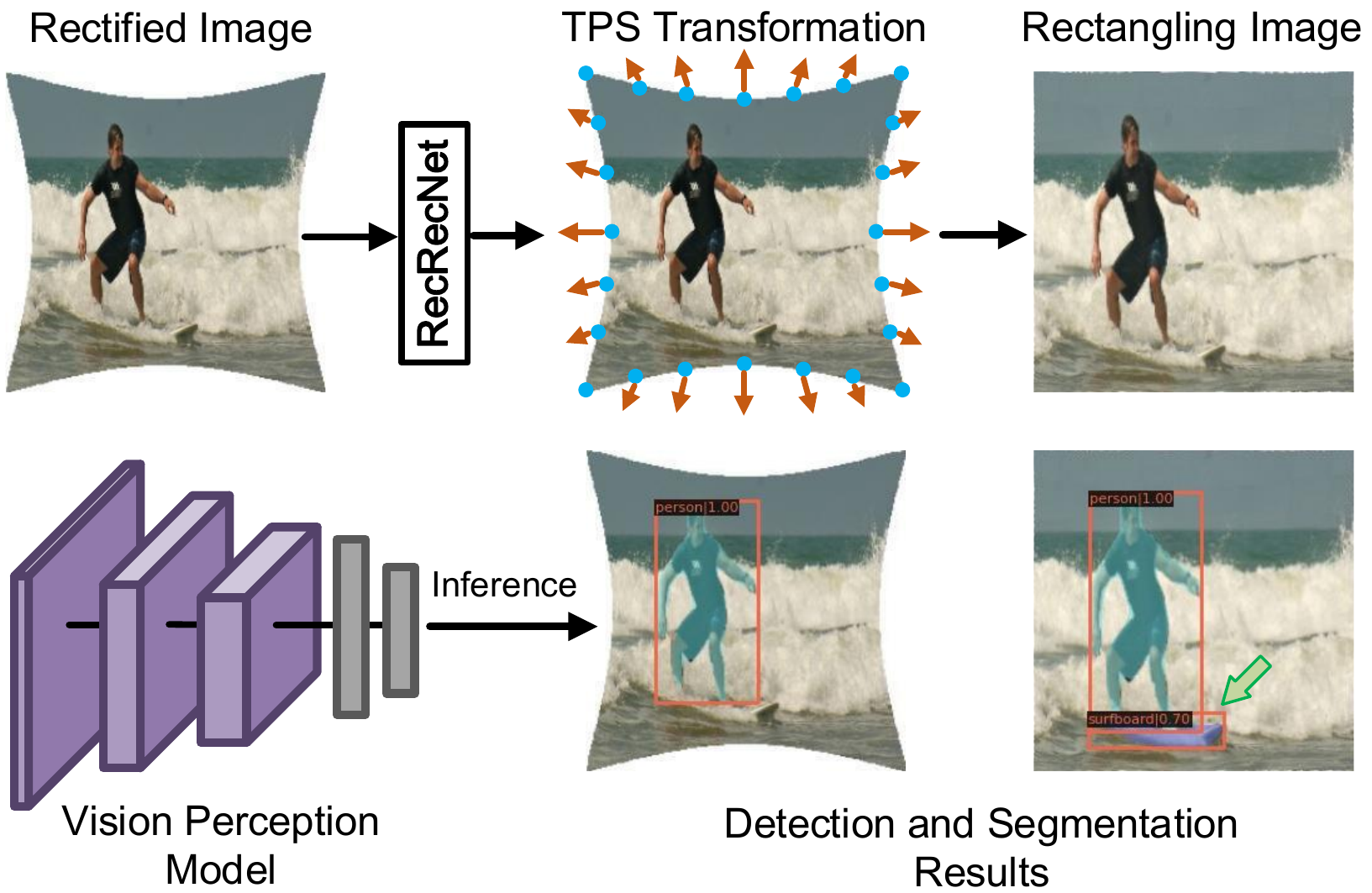}
\caption{Existing rectification methods devote to correcting the content of the wide-angle image, but they inevitably distort the image boundary. Such a deformed boundary can lead to dramatic performance degradation of the popular vision perception model~\cite{he2017mask}. By presenting RecRecNet, we can reach a win-win representation on both the undistorted content and regular boundary. The arrow marks our perception gain (\textit{e.g.}, the surfboard is rediscovered).}
\label{fig:teaser}
\vspace{-0.4cm}
\end{figure}

\section{Introduction}
\label{sec:introduction}

The wide-angle lens has been widely applied in various applications such as computational imaging, virtual reality (VR), and autonomous driving, due to their larger field of view (FoV) than conventional lenses~\cite{kannala2006generic, gao2022review, yang2021capturing, liao2023deep}. By employing a specific mapping, the wide-angle lens can produce convex and non-rectilinear images rather than images with straight lines of perspective. However, this mapping mechanism violates the pinhole camera assumption, and the captured image suffers from severe radial distortions.

To eliminate the distortion induced by the wide-angle lens, the distortion rectification algorithm is presented. For example, the traditional approaches calibrate the wide-angle lens using the calibration target (\textit{e.g.}, a checkerboard)~\cite{zhang2000flexible, kannala2006generic} and specific scene assumption~\cite{17, 18}. In recent years, deep learning has brought new inspirations to this field. It helps to automatically rectify the radial distortion in wide-angle images based on learned semantic features. To pursue better performance, the recent works improve the learning framework through diverse features~\cite{FishEyeRecNet, Xue}, learning paradigm~\cite{DR-GAN}, calibration representation~\cite{BlindCor, Lopez}, and distortion prior~\cite{FE-GAN, OrdianlDistortion}, etc. 

We notice that the above distortion rectification approaches mainly focus on rectifying the image content. By shrinking the distorted image towards the principal center, the spatial distribution of all pixels has been rearranged and thus generates the rectified result. However, the image boundaries are inevitably distorted by this correction way, making the image exhibited with a visually narrow FoV and irregular shape. Moreover, we found such a deformed boundary can significantly influence the current mainstream vision perception models because most of them are trained on the dataset with rectangular images. As shown in Figure~\ref{fig:teaser}, Mask R-CNN~\cite{he2017mask} is confused by the rectified image, leading to a missing segmentation and detection result near the boundary (the surfboard is neglected ). While some works~\cite{liao2021towards} fill the blank region beyond deformed boundaries using outpainting strategy~\cite{111, teterwak2019boundless}, the generated contents are usually fictitious and twist the original semantic characterization.

In this work, we explore a win-win rectification representation for the wide-angle image on \textit{both} content and boundary without introducing any extra semantic information~\cite{he2013rectangling, nie2022deep}. For this new representation, the distorted content has been corrected while the image boundary keeps straight, as shown in Figure~\ref{fig:teaser}. To this end, we propose a \uline{Rec}tangling \uline{Rec}tification \uline{Net}work (named \textbf{RecRecNet}). Concretely, a thin-plate spline (TPS) motion module is proposed to formulate the non-linear and non-rigid transformation for rectangling the rectified wide-angle image. It can flexibly warp the source structure to the target domain by learning the control points on the rectified image. The control points are located at the whole distribution of an image, while Figure~\ref{fig:teaser} only shows a part of the points for intuitive clarification.

Subsequently, we design a DoF-based curriculum learning to further inspire our RecRecNet to grasp the progressive deformation rules, relieving the burden of complex structure approximation and initializing a better start point for training. The general paradigm of curriculum learning~\cite{Bengio} introduces a heuristic and purposeful strategy for training a learning model. The model can converge faster by gradually learning various samples or knowledge based on different stages in a curriculum, mimicking how humans and animals learn. In this regard, we increase the degree of freedom (DoF) of the devised stages/transformations from similarity transformation (4-DoF) to homography transformation (8-DoF). Moreover, the procedure of our curriculum reveals a simple-to-complex order from the straight line, sloped line to curve for the image boundary. As a consequence, our RecRecNet can discover more geometric details and converge faster on the final rectangling stage.

While aiming for a win-win rectification representation, we also provide an in-depth analysis of why the deformed boundary makes the perception deformed, \textit{i.e.}, why the deformed image boundary can significantly influence the visual perception models. We surprisingly found that the deformed boundary can introduce new features on the original feature maps, which further forms new semantics or cause blind spots for non-salient features. And we demonstrate that this effect commonly exists in other research regions.

We conclude our major contributions as follows:
\begin{itemize}
    \item We propose a Rectangling Rectification Network (RecRecNet) for a win-win representation of the rectified wide-angle image. To our knowledge, it has never been investigated in previous literature.
    \item A thin-plate spline (TPS) motion module is proposed to flexibly formulate the non-linear and non-rigid rectangling transformation. We also provide an in-depth analysis of why the deformed image boundary can significantly influence the visual perception models.
    \item To grasp the progressive deformation rules and relieve the burden of complex structure approximation, a DoF-based curriculum learning is designed to inspire our RecRecNet.
 \end{itemize}
 
\begin{figure*}[t]
\centering
\includegraphics[width=1\linewidth]{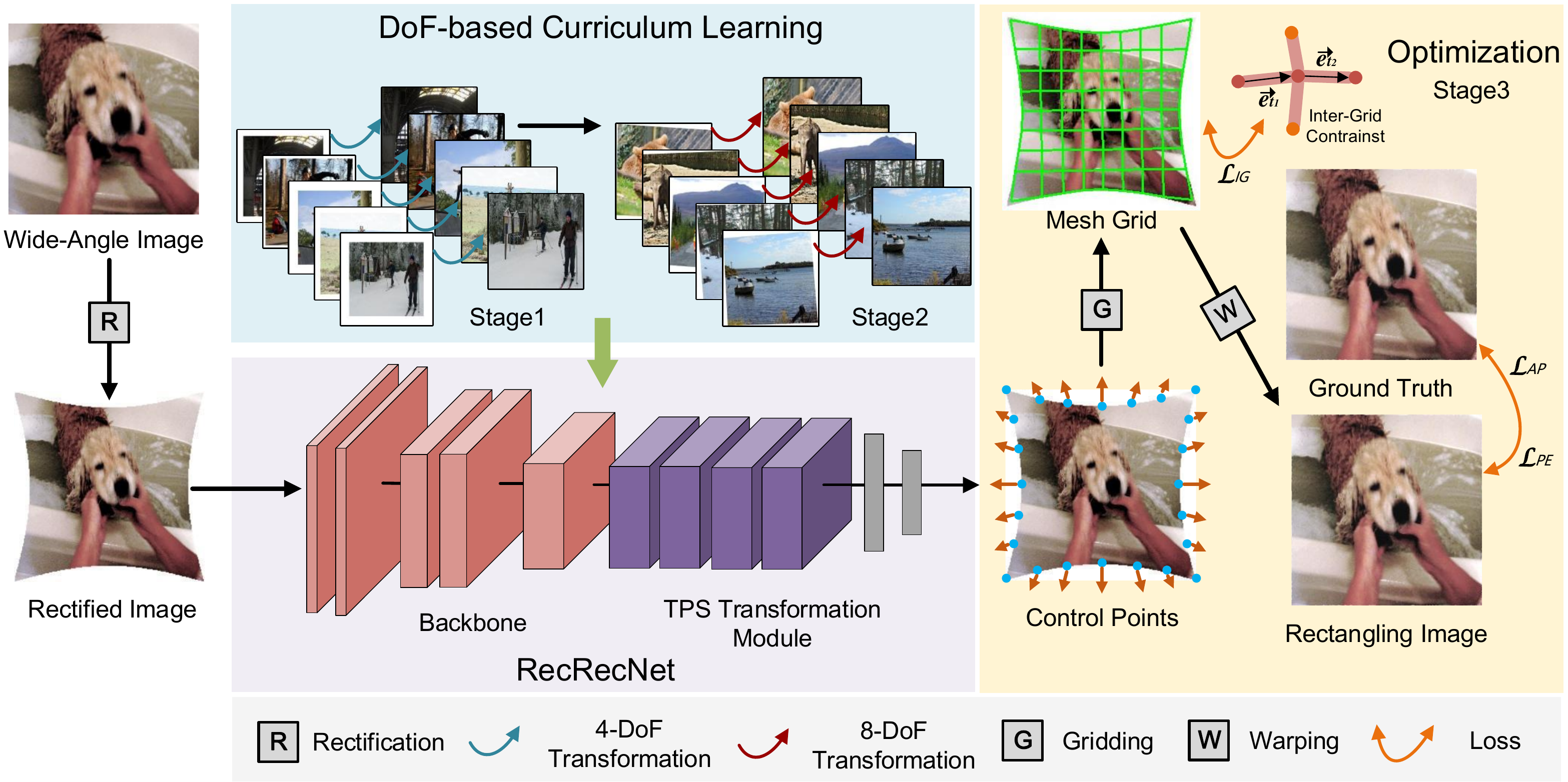}
\caption{Overview of the proposed framework. Our Rectangling Rectification Network (RecRecNet) aims to construct a win-win representation on both image content and image boundary. RecRecNet learns a flexible local transformation based on the thin-plate spline (TPS) motion model, which predicts the control points on the image boundary and warps the target image by using a mesh grid. It is jointly optimized with the appearance constraint $\mathcal{L}_{AP}$, perceptual constraint $\mathcal{L}_{PE}$, and inter-grid constraint $\mathcal{L}_{IG}$. To further relieve the challenges of structure approximation, we inspire our RecRecNet to learn the gradual deformation rules with a DoF-based curriculum learning.}
\vspace{-0.3cm}
\label{fig:framework}
\end{figure*}

\section{Related Work}
\label{sec:related_work}

\subsection{Distortion Rectification}
The distortion rectification approaches for wide-angle lenses can be classified into traditional and learning-based approaches. The majority of traditional techniques concentrate on detecting the hand-crafted features \cite{19,21, 17, 18}. However, these techniques frequently fail due to particular constraints like the plumb line and curve. Recently, blind distortion rectification has been studied using deep learning. It can rectify the radial distortion of the wide-angle image without any scene assumptions. For example, the regression-based solutions \cite{Rong, DeepCalib, FishEyeRecNet, Lopez, OrdianlDistortion} predict the distortion parameters of input, and then the estimated parameters are used to rectify the distortion offline or online. The reconstruction-based solutions \cite{DR-GAN, BlindCor, PCN, FE-GAN} directly learn the pixel-level mapping or displacement field from the distortion image into the rectification output.

Despite promising flexibility and efficiency, these rectification approaches inevitably distort the image boundary during content correction. As a result, the distorted distribution at the boundary confuses the vision perception network and leads to inferior performance. One direct way is to crop the rectified image into a rectangular shape~\cite{he2013content}, however, it discards informative content and disobeys the original intention of the wide-angle lens.

\subsection{Curriculum Learning}
Curriculum learning techniques have been effectively exploited across the board in machine learning. Elman et al.~\cite{Elman} provide a conceptual framework for curriculum learning that emphasizes the value of beginning at a low level and then works way up to more challenging cases. Bengio et al.~\cite{Bengio} are the first to formalize curriculum learning as a strategy to gradually raise the level of complexity of the data samples utilized in the training process. Subsequently, most researchers follow this paradigm and apply the curriculum learning into numerous regions~\cite{shi2015recurrent, tudor2016hard, pentina2015curriculum}. Particularly in computer vision, curriculum learning has been investigated to inspire the learning networks on image classification~\cite{cascante2021curriculum, morerio2017curriculum, guo2018curriculumnet}, object detection~\cite{li2017multiple, wang2018weakly, soviany2021curriculum}, and semantic segmentation~\cite{zhang2017curriculum, sakaridis2019guided, feng2020semi}, etc.

In this work, we incorporate curriculum learning into a new task. By disassembling the rectangling rectification into three levels based on the DoF of transformation, the proposed curriculum can inspire our network to learn the challenging deformation in a simple-to-complex order and thus facilitates a faster convergence for training.

\section{Methodology}
\label{sec:methodology}
\subsection{Problem Formulation}
Given a rectified wide-angle image $I_{Rec} \in \mathbb{R}^{h\times w\times 3}$ with deformed boundaries, our Rectangling Rectification Network (RecRecNet) aims to construct a win-win representation on both image content and image boundary, generating a rectangular rectified image $I_{{Rec}^2} \in \mathbb{R}^{h\times w\times 3}$. Figure~\ref{fig:framework} shows the overview of the proposed framework. To be more specific, the rectangling module takes $I_{Rec}$ as the input and learns a flexible local transformation $\mathcal{T}$ based on the thin-plate spline (TPS) motion model~\cite{tps, jaderberg2015spatial, liu2018thin, detone2018cvpr}. Compared with prior TPS works, our method aims to formulate the nonlinear and non-rigid rectangling transformation. The accurate localization of control points on curved image boundaries is the biggest challenge. With the appearance constraint and inter-grid constraint, we can learn the control points with fewer distortions in an end-to-end unsupervised manner. To further relieve the challenges of structure approximation, we inspire the network to learn the gradual deformation rules with a DoF-based curriculum learning.

The general polynomial camera model~\cite{kannala2006generic} is widely used in the approximation for the radial distortion of the wide-angle image. We formulate the relationship between the lens's projection manner and its distortion parameters $\mathcal{K} = [k_1, k_2, k_3, \cdots]$ as follows:

\begin{equation}\label{eq_radial_distortion}
r(\theta) = \sum_{i = 1}^{N}{k_i \theta^{2i - 1}}, N = 1, 2, 3, \cdots,
\end{equation}
where $r$ is the distance between the principal point and the pixels in the image, and $\theta$ indicates the angle between the optical axis and incident ray. Such a projection can produce a convex and non-rectilinear image, and thus the captured image suffers from severe geometric distortions. Suppose that the coordinates of a pixel in the wide-angle image and rectified images are $\bm{p}=(x,y)$ and $\bm{p}_{rec}=(x_{rec},y_{rec})$, the rectification solution can be then described as follows:
\begin{equation}\label{eq:rectification}
    \bm{p}_d = \mathcal{R}(\bm{p}_{rec},\mathcal{K})=\begin{pmatrix}
    u_0\\v_0
    \end{pmatrix} + \frac{r(\theta)\bm{p}_{rec}}{\left\|\bm{p}_{rec}\right\|_2},
\end{equation}
where $(u_0, v_0)$ represents the principal point. With the above equation, the wide-angle image can be rectified by using bilinear interpolation. By shrinking the wide-angle image towards the principal center, the spatial distribution of all pixels is rearranged, and thus rectified content is obtained. However, the image boundaries are inevitably distorted by this rectification way.

\subsection{Rectangling with TPS Transformation}
As shown in Eq.~\ref{eq:rectification}, the rectification of the wide-angle image is a non-linear and non-rigid mapping function, which cannot be represented by simple transformations. Besides, it is challenging to build an accurate parameterization model to supervise the rectangling process with estimated parameters. From the perspective of energy minimization, we can approximate the rectangling rectification by minimizing the transformation energy with a uniform model:
\begin{equation}
    \begin{split}
        \varepsilon = \varepsilon_{\mathcal{T}} &+ \lambda \varepsilon_{d},\\
        \mathcal{T} : \, (x_{rec},y_{rec}) &\mapsto  (x_{{rec}^{2}},y_{{rec}^{2}}),\\
    \end{split}
    \label{eq:deformation-model}
\end{equation}
where $\varepsilon$ denotes the total energy of the expected transformation. $(x_{rec},y_{rec}) \in \Omega_S$ and $(x_{{rec}^{2}},y_{{rec}^{2}}) \in \Omega_T$ represent the point in the domain of source $S$ and target $T$, respectively. $\varepsilon_{\mathcal{T}}$ indicate the data penalty energy and $\varepsilon_{d}$ indicate the distortion energy. The above formula with the lowest overall energy is the desired transformation, in which a hyperparameter $\lambda$ is designed to balance the energies between the data penalty and distortion. 

Particularly, we propose to exploit TPS transformation $\mathcal{T}'$ to minimize the aforementioned energy function, which enables an end-to-end unsupervised deformation. TPS transformation provides for the representation of more complex motions since it is flexible and non-linear. It allows warping from one image to the other with minimum distortion, given two sets of control points in the corresponding images. Suppose $\bm{q} = [q_1, q_2, \cdots, q_N]$ and $\bm{q}' = [q'_1, q'_2, \cdots, q'_N]$ are the source points and the target points, the data term $\varepsilon_{\mathcal{T}'}$ can be calculated by the difference of control points displacement $\sum_{i=1}^{N} \lVert \mathcal{T}'({\bm{q}_{i}})- {\bm{q}'_{i}}\rVert_{2}^{2}$. Having aligned the control points, TPS finds an optimal interpolation transformation with minimal distortion term $\varepsilon_{d}$ by:
\begin{equation}
  \label{eq_tps}
  \begin{matrix}
    \begin{aligned}
      \min \iint_{\mathbb{R}^{2}}\left(\left(\frac{\partial^{2} \mathcal{T}'}{\partial x^{2}}\right)^{2}\right.+2\left(\frac{\partial^{2} \mathcal{T}'}{\partial x \partial y}\right)^{2} \\ +\left. \left(\frac{\partial^{2} \mathcal{T}'}{\partial y^{2}}\right)^{2}\right) dx dy.
    \end{aligned}\\ 
  \end{matrix}
\end{equation}
The above equation utilizes second-order derivatives to formulate the distortion deviation of each target point and constrains a cumulative global minimum. Then we can reach a spatial deformation function parameterized by control points as follows:
\begin{equation}
\label{eq_tps_cp}
  \mathcal{T}'(q)=A\begin{bmatrix}
    q\\ 
    1   
    \end{bmatrix}+\sum_{i=1}^{N} w_{i} U\left({\left\lVert \bm{q}'_{i}-q\right\rVert}_2\right),
\end{equation}
where $q$ is a point located at the rectified wide-angle image. $A \in \mathbb{R}^{2\times 3}$ and $w_{i} \in \mathbb{R}^{2\times 1}$ are the transformation parameters~\cite{tps}, which can be derived by minimizing Eq.~\ref{eq:deformation-model}. Besides, $U(r)$ is a radial basis function that denotes the influence of the control point on $\bm{q}$: $U(r)=r^{2} \log r^{2}$.

In RecRecNet, we define $N = (U + 1)\times(V + 1)$ control points that can be connected to form a mesh, and the network learns to predict the mesh of rectified wide-angle image. Then TPS transforms the predicted mesh to the regular mesh on the target rectangular image, where the control points are set to be evenly distributed on the image.

For the network, we first choose ResNet50~\cite{resnet} as the backbone of our RecRecNet to extract the high-level semantic features. Suppose the size of the input rectified image is $W\times H$, and the size of the output feature map of the backbone is $\frac{W}{16}\times \frac{H}{16}$. Then these feature maps are fed to a motion header to predict $(U + 1)\times(V + 1)$ control points on the rectified image. 4 convolutional layers with a kernel size of 3 and BacthNormalization are stacked to aggregate the high-level features, of which all channel dimensions are set to 512. Subsequently, 3 fully connected layers with the following numbers of units 4096, 2048, $(U + 1)\times(V + 1)\times2$ are used to predict the coordinates of all control points.

\subsection{DoF-based Curriculum Learning}
As mentioned above, we propose to exploit TPS transformation to flexibly approximate the motion from the rectified image to rectangling result by a set of control points. To further relieve the challenges of this transformation, we inspire our RecRecNet with a curriculum learning strategy.

\subsubsection{Image Transformation Formulation}
\label{sec_trans_formulation}
Image transformation contains different elements such as translations, scales, rotations, shearing, etc. Any composition of these elements can form a new transformation with the degree of freedom (DoF). In general, the degree of freedom is a collection of independent displacements that completely describes a system's displaced or deformed position. There are some classical transformations on 2D images: translation transformation (2-DoF), Euclidean/Rigid transformation (translation + rotation, 3-DoF), similarity transformation (translation + rotation + scale, 4-DoF), affine transformation (translation + rotation + scale + shear, 6-DoF), and projective transformation (two each for translation, rotation, scale, and lines at infinity, 8-DoF).

\begin{figure}[t]
\centering
\includegraphics[width=1\linewidth]{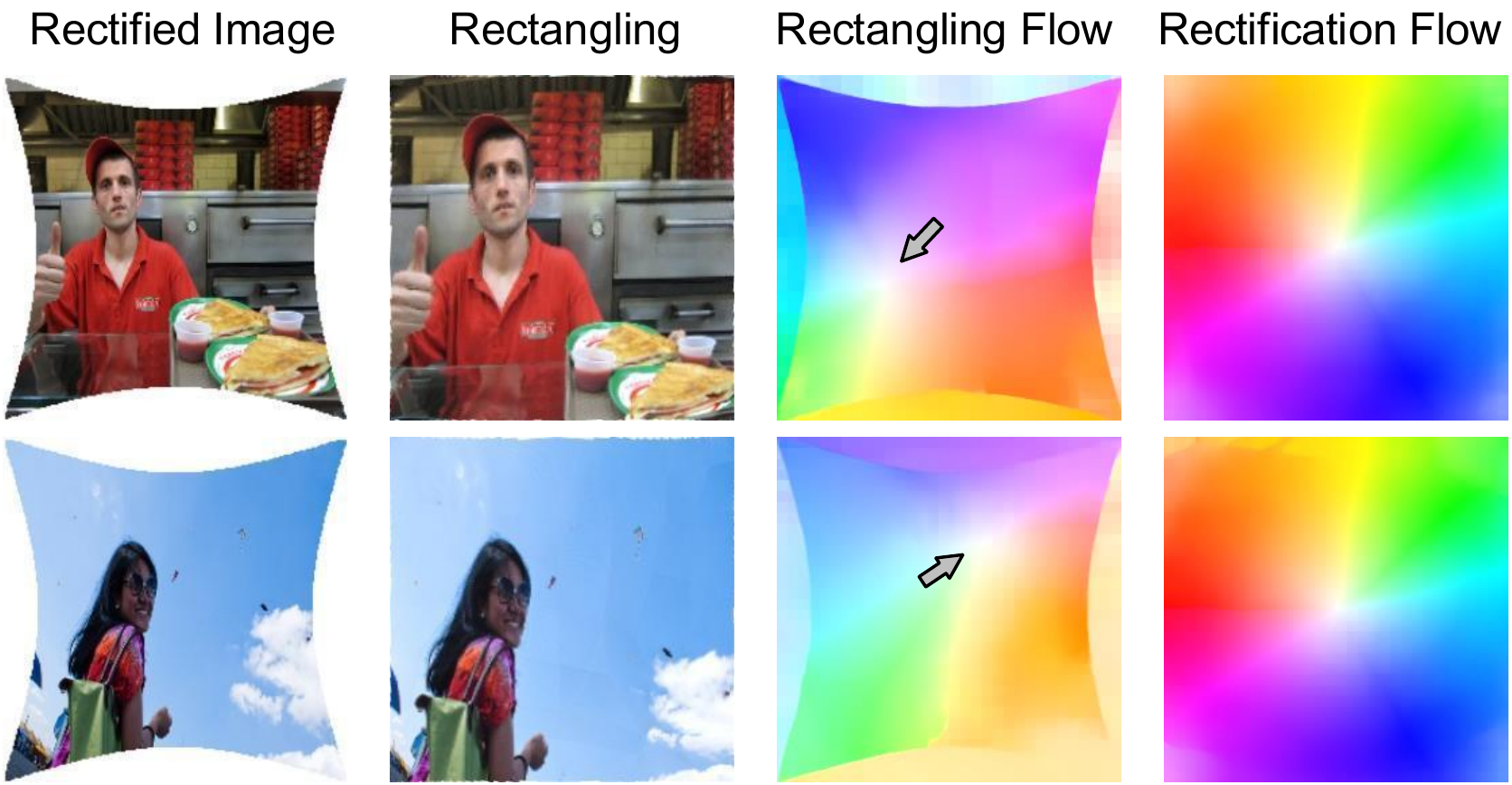}
\caption{Visualization of the rectangling flow and rectification flow for the rectified wide-angle image. We mark the different radial centers in different rectangling flows with arrows.}
\vspace{-0.4cm}
\label{fig:flow}
\end{figure}

The image transformation between the rectified wide-angle image and its rectangling image has never been studied and formulated. Compared to distortion rectification, the rectangling transformation seems to be an inverse asymmetric warping due to the more important image boundary. To understand this transformation intuitively, we visualize the pixel displacement field between the rectified wide-angle image and its rectangling image. Specifically, we leverage the state-of-the-art optical flow estimation network RAFT~\cite{teed2020raft} to exhibit the motion difference and name it as rectangling flow (rectified image $\mapsto$ rectangling image). As shown in Figure~\ref{fig:flow}, compared with the rectification flow (wide-angle image $\mapsto$ rectified image), we have the following observations: (1) Both rectangling flow and rectification flow show a radial spatial distribution, in which four dominated directions spread towards or backward four image corners. (2) Rectangling flow has an unfixed radial center for each sample, which is determined by the content and layout of the image. Instead, the rectification flow has a relatively fixed radial center, which is related to the geometry prior to the radial distortion in the wide-angle image.

Thus, we can conclude the image content highly identifies the distribution of rectangling flow. It is painful to represent this transformation using classical image transformations and their compositions. Nevertheless, such a non-linear and non-rigid transformation can be locally disentangled into different transformation elements such as translation, scale, and shearing. It would be meaningful to inspire the network to learn some basic transformations and gradually transfer them into more challenging cases.

\subsubsection{Curriculum Selection}
To choose effective curriculums for the rectangling task, we should obey two principles: First, all stages in a curriculum have closely associated knowledge of the target. Second, the stages in a curriculum intrinsically offer a simple-to-complex order. To this end, we design a DoF-based curriculum to inspire RecRecNet, of which we increase the DoF of the devised transformations from similarity transformation (4-DoF) to homography transformation (8-DoF) and eventually to the rectangling transformation.

We argue that the proposed curriculum has the following strengths: (1) As mentioned in Section~\ref{sec_trans_formulation}, the DoF directly expresses the difficulty of the image transformation. As the DoF increases, the transformation shows more complex global mappings and diverse motions. Thus, RecRecNet can learn the gradual deformation rules. (2) For the local geometry, \textit{i.e.}, the shape of the image boundary, our curriculum shows a progressive shift from a straight line, sloped line, to a curve. Such a design can improve the localization ability of control points in TPS transformation and enable a boundary-aware ability for RecRecNet. More details and experiments will be demonstrated in Section~\ref{sec_ablation}.

\begin{figure*}[t]
\centering
\includegraphics[width=1\linewidth]{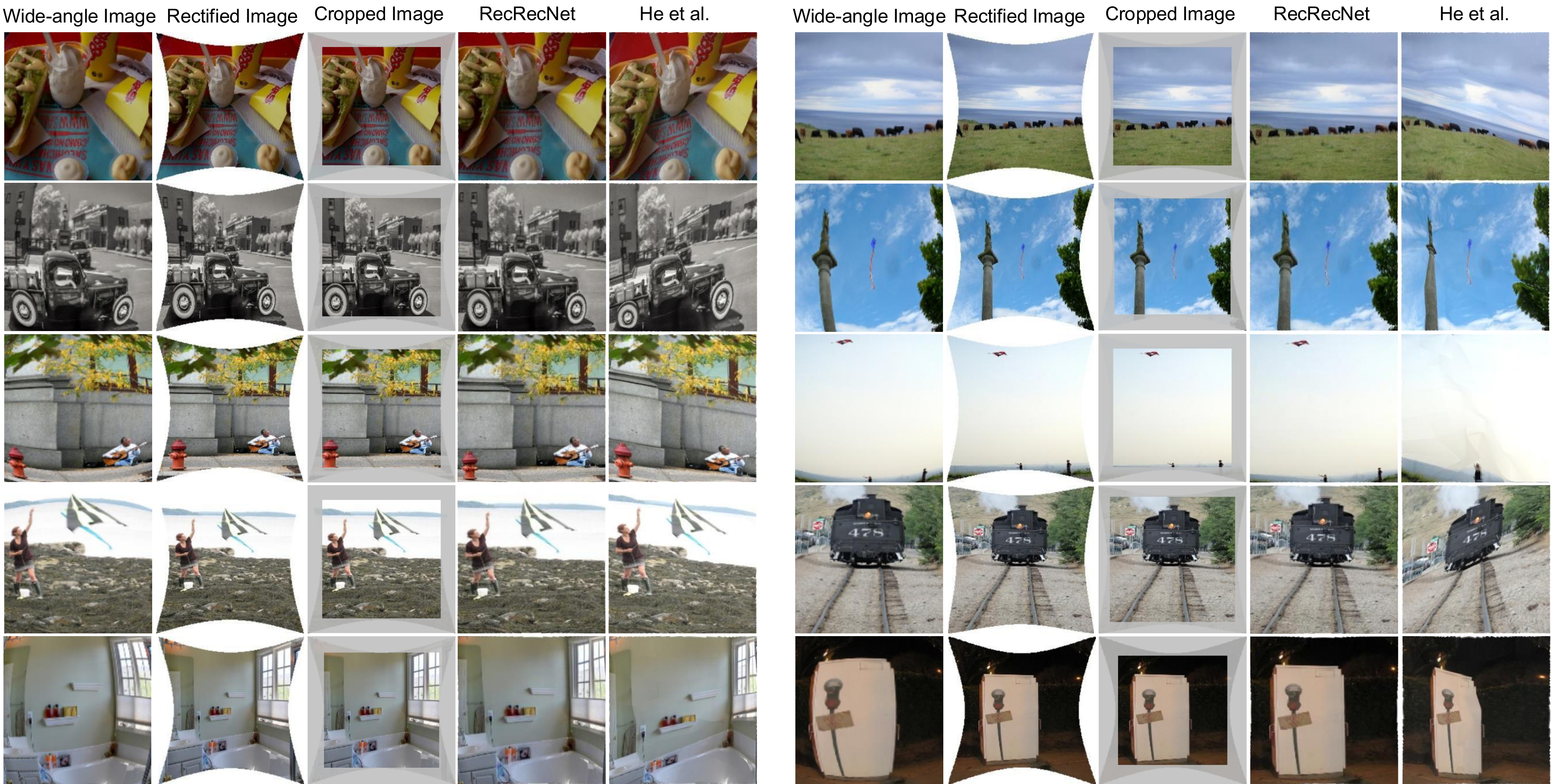}
\caption{Comparison to the image rectangling work He et al.~\cite{he2013rectangling}. For a comprehensive evaluation, we split the test dataset into the complex scene (left) and the simple scene (right).}
\vspace{-0.35cm}
\label{fig:cp_he}
\end{figure*}

\begin{figure}[t]
\centering
\includegraphics[width=1\linewidth]{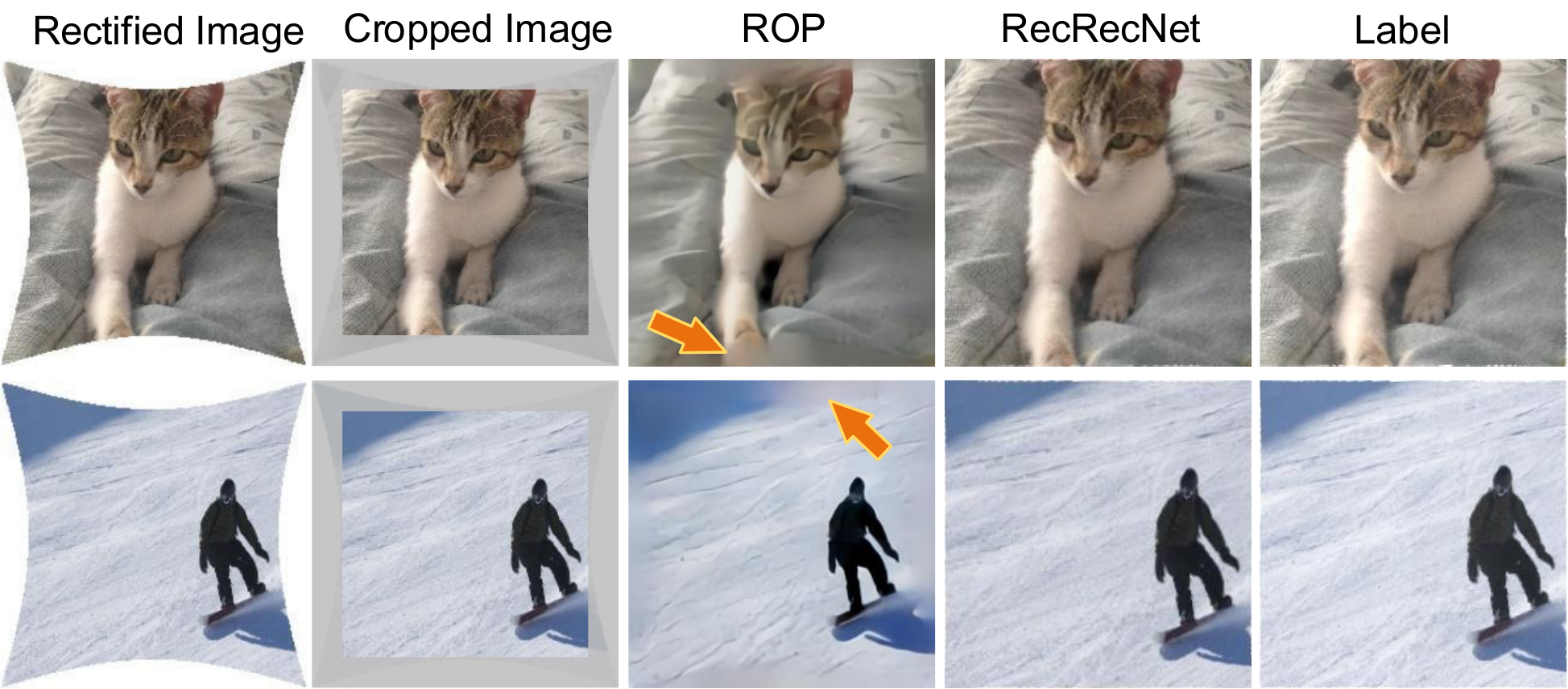}
\caption{Comparison to the rectification outpainting work ROP~\cite{liao2021towards}. Our RecRecNet can straighten the irregular image boundary in the rectified image without introducing new content.}
\vspace{-0.5cm}
\label{fig:cp_rop}
\end{figure}

\subsection{Training Losses}

\noindent\textbf{Appearance Loss:} 
We constrain the rectangling image $I_{{Rec}^2}$ to be close to the ground truth $I_{GT}$ at the pixel level. The appearance loss $\mathcal{L}_{AP}$ can be calculated by the difference between $I_{{Rec}^2}$ and $I_{GT}$ with $\mathcal{L}_1$ norm.

\noindent\textbf{Perceptual Loss:} We then minimize the $\mathcal{L}_2$ distance between $I_{{Rec}^2}$ and $I_{GT}$ in high-level semantic perception to guarantee the rectangling results perceptually natural:
\begin{equation}\label{eq_loss_pe}
\begin{split}
\mathcal{L}_{PE} = \frac{1}{W_{i,j}H_{i,j}} \sum_{x=1}^{W_{i,j}}\sum_{y=1}^{H_{i,j}}||\phi_{i,j}{(I_{{Rec}^2}^{x,y})} - \phi_{i,j}{(I_{GT}^{x,y})}||_2, 
\end{split}
\end{equation}
where the difference of rectangling rectified image and ground truth is minimized on the feature map $\phi_{i,j}$, which is derived from the $j$-th convolution (after activation) before the $i$-th max-pooling layer in the VGG19 network.

\noindent\textbf{Inter-grid Mesh Loss:} To avoid the content distortion after our rectangling, the predicted mesh should not be largely deformed. Therefore, we design an inter-grid mesh term $\mathcal{L}_{IG}$ to constrain the shape of the predicted mesh, which encourages the neighboring grids to transform consistently. The edges of two successive deformed grid $\{\vec{e}_{t1}, \vec{e}_{t2}\}$ are supervised to be co-linear as follows:
\begin{equation}\label{eq_loss_ig}
  \mathcal{L}_{IG}= \frac{1}{M}\sum_{\{\vec{e}_{t1}, \vec{e}_{t2}\}\in m_s}(1-\frac{\langle \vec{e}_{t1},\vec{e}_{t2}\rangle}{\parallel \vec{e}_{t1}\parallel \cdot \parallel \vec{e}_{t2}\parallel }),
\end{equation}
where $M$ is the number of tuples of two successive edges in a mesh $m_s$. When maximizing the above cosine representation, \textit{i.e.}, the value equals 1, and the corresponding two edges are co-linear. Thus, the loss reaches the minimum and the image content exhibits consistently.

The overall training loss can be obtained by:
\begin{equation}\label{eq_loss_all}
\mathcal{L} = \lambda_{AP}\mathcal{L}_{AP} + \lambda_{PE}\mathcal{L}_{PE} + \lambda_{IG}\mathcal{L}_{IG},
\end{equation}
where $\lambda_{AP}$, $\lambda_{PE}$, and $\lambda_{IG}$ are the weights to balance the appearance loss, perceptual loss, and inter-grid mesh loss, which are empirically set to $1$, $1e^{-4}$, and $1$.

\section{Experiments}
\label{sec:experiment}
\subsection{Implementation Details}
\noindent\textbf{Dataset Establishment:} 
Following existing methods~\cite{Rong, FishEyeRecNet, DR-GAN, PCN}, we first synthesize the wide-angle images by using a $4^{th}$ order polynomial model based on Eq.~\ref{eq_radial_distortion}. Then we perform the distortion rectification on the wide-angle image in terms of Eq.~\ref{eq:rectification}. Due to non-linear and non-rigid characteristics, forming an accurate transformation model for rectangling the rectified image is difficult. We notice that there is a classical panoramic image rectangling work by He et al.~\cite{he2013rectangling} on computer graphics. It makes the stitched image regular by optimizing an energy function with line-preserving mesh deformation. Thus, we perform the same energy function on the rectified image to fit our task. However, the capability to preserve linear structures ~\cite{he2013rectangling} is limited by line detection. Consequently, some rectangling images have non-negligible distortions. To overcome this issue, we carefully filter all results and repeat the selection process three times, resulting in 5,160 training data from 30,000 source images and 500 test data from 2,000 source images. Each manual operation takes around 10s. More details about the dataset are reported in the supplementary material. We would like to release the dataset to promote the development of the related community.

\noindent\textbf{Experimental Configuration:}
We train RecRecNet with an exponentially decaying learning rate with an initial value of $10^{-4}$ using Adam\cite{kingma2014adam}. The batch size is set to 16 and the total epoch is set to 260. Particularly, RecRecNet is trained with 3 stages in terms of the proposed DoF-based curriculum. The training epochs of the 4-DoF curriculum, 8-DoF curriculum, and final rectangling curriculum are empirically divided into 30, 50, and 180.

\subsection{Rectangling Rectification Results}
\begin{figure}[t]
\centering
\includegraphics[width=.95\linewidth]{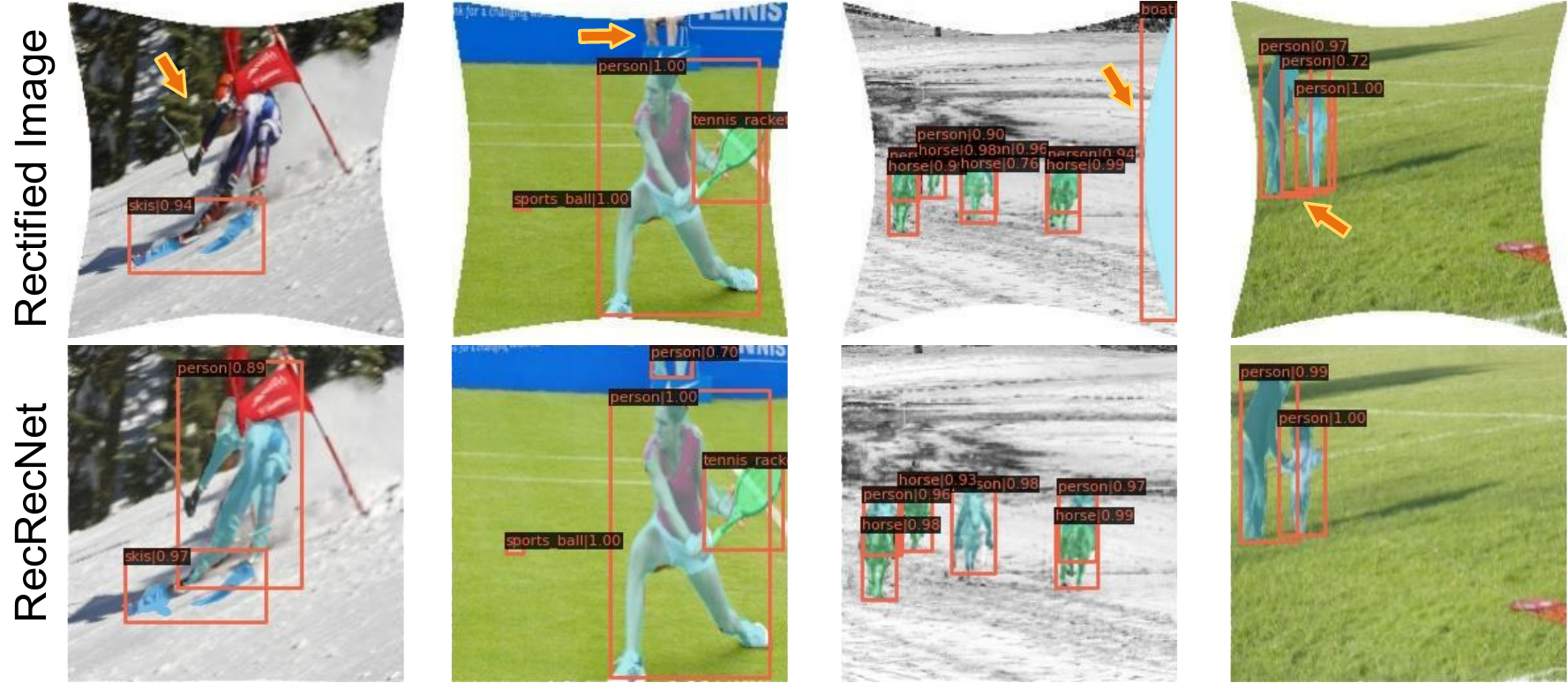}
\caption{RecRecNet helps the downstream vision perception tasks. The arrows mark the failure results in rectified images.}
\vspace{-0.4cm}
\label{fig:perception_results}
\end{figure}

RecRecNet is the first effort to rectangling the rectified wide-angle image. We compare our method with the classical image rectangling work He et al.~\cite{he2013rectangling} and the state-of-art outpainting work for rectified image~\cite{liao2021towards}. In the first comparison (Figure~\ref{fig:cp_he}), we split the test dataset into the complex scene and simple scene based on the image feature. We also show the cropped result that directly discards the content around the deformed boundary, which is commonly applied in previous works. Experimental results demonstrate while RecRecNet is trained based on the filtered labels from He et al.~\cite{he2013rectangling}, it can learn a higher upper bound of performance and generalize to more diverse scenes. By contrast, He et al.~\cite{he2013rectangling} fail to rectangling the scene with a complicated layout or monotonous background due to line-preserving rules. In the second comparison (Figure~\ref{fig:cp_rop}), we can observe ROP~\cite{liao2021towards} makes the deformed boundary of the rectified image regular by extrapolating the image content. But the generated results are usually fictitious and twist the original semantic characterization, in which an unpleasant fracture of semantics occurs near the boundary. Without introducing new and ambiguous content, our RecRecNet constructs a win-win representation for the rectified image using a flexible TPS transformation, showing the closest appearances to the rectangling label. Besides, Figure~\ref{fig:perception_results} shows RecRecNet can considerably help the downstream vision tasks.

\begin{table}[h]
\vspace{-0.2cm}
\begin{center}
\caption{Quantitative evaluation on comparison methods.}
\label{tab:quantitative}
\footnotesize
\begin{tabular}{l|ccccc}
\hline
Metrics & Crop & Padding & ROP~\cite{liao2021towards} & He~\cite{he2013rectangling} & Ours\\
\hline
\hline
PSNR $\uparrow$ & 11.51 & 12.07 & 13.90 &15.36 &18.68\\
SSIM $\uparrow$ & 0.1907 & 0.2775 & 0.3516 &0.4211 &0.5450\\
\hline
AP $\uparrow$ & 21.8 & 23.9 & 30.8 &34.7 &41.3\\
mIoU $\uparrow$ & 18.6 & 20.1 & 26.1 &30.2 &37.8 \\
\hline
\end{tabular}
\end{center}
\vspace{-0.5cm}
\end{table}

Furthermore, we compare different methods based on quantitative evaluation from two aspects. For the rectangling results, Peak Signal-to-Noise Ratio (PSNR) and Structural Similarity (SSIM) are selected to measure the image quality. For vision tasks, we leverage the Average Precision (AP) and Mean Intersection over Union (mIoU) to evaluate the detection and segmentation performance of the vision model~\cite{he2017mask}. The comparisons are listed in Table~\ref{tab:quantitative}, where we leverage the mirror padding as the Padding method to simply fill the blank region beyond the deformed boundary. As we can observe, the evaluation results show the superiority of RecRecNet not only in low-level visual reconstruction but also in high-level semantic recovery.

\begin{figure}[t]
\centering
\includegraphics[width=.95\linewidth]{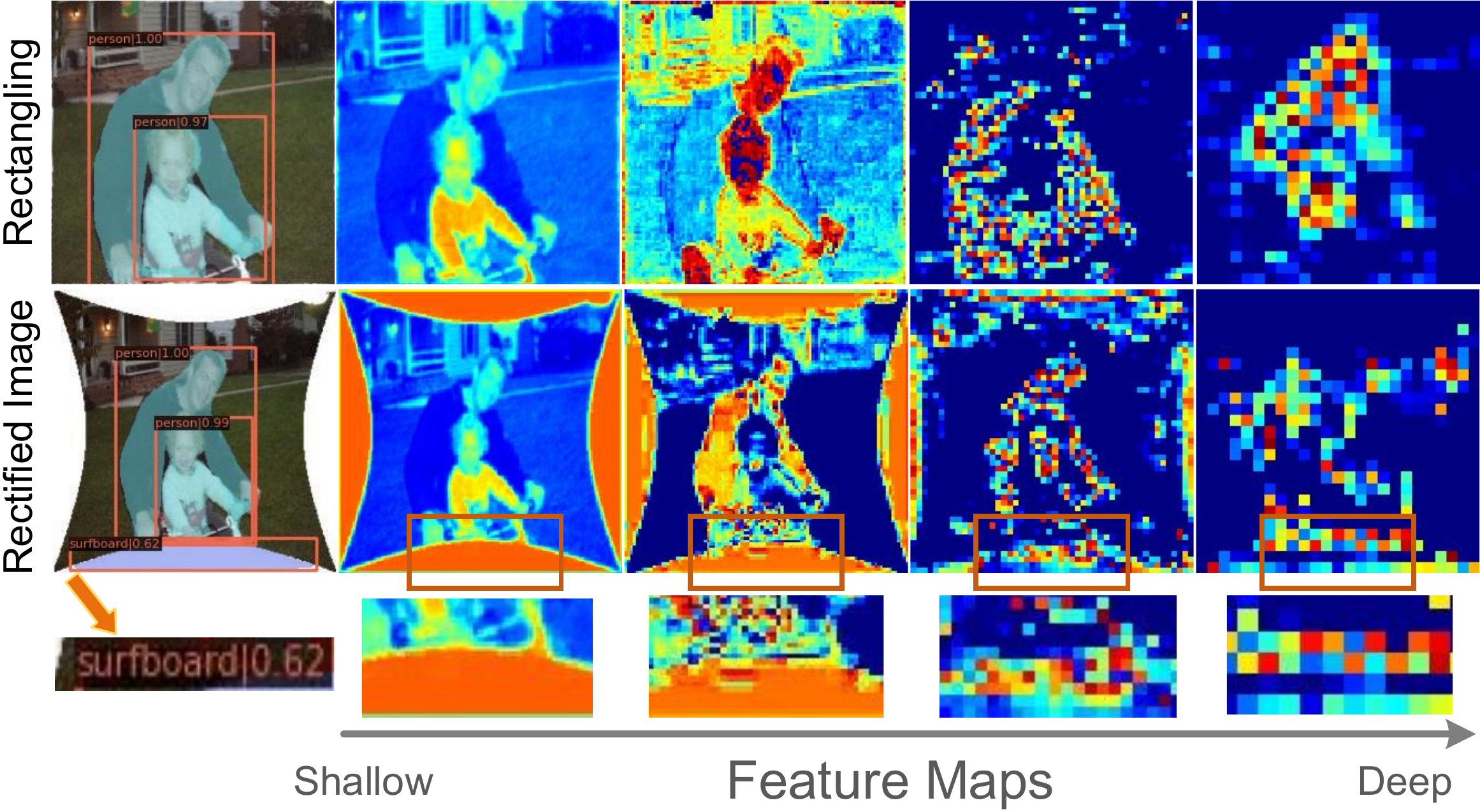}
\caption{Failure case of Mask R-CNN~\cite{he2017mask} in regards to the wrong perception. We find that the deformed boundary can introduce new features to the original feature maps.}
\vspace{-0.3cm}
\label{fig:wrong_perception}
\end{figure}

\subsection{Why Deformed Boundary Makes Perception Deformed}
While our task is to rectangling the rectified image, we are still curious about the secret behind the performance degradation of the vision perception models. In this part, we provide an in-depth analysis of this effect. 
First, two typical failure cases of the perception model are formulated: wrong perception and missing perception. Concretely, we focus on the detection performance of Mask R-CNN~\cite{he2017mask} and visualize some samples in Figure~\ref{fig:perception_results} and Figure \ref{fig:wrong_perception}. We can observe the model fails to detect the objects, especially those located near the boundary. By rectangling the deformed boundary, the perception performance is surprisingly restored.

In experiments, we found the deformed boundary can introduce \textit{new features} onto the feature maps, which (i) form new semantics (leads to the wrong perception) or (ii) cause blind spots for original features (leads to missing perception). Especially in Figure \ref{fig:wrong_perception}, we exhibit the feature maps of Mask R-CNN with ResNet101~\cite{resnet} backbone from shallow layers to deep layers. The noticeable line artifacts and curve artifacts can be observed at the outermost boundaries and deformed boundaries, respectively. In particular, the line artifacts are essentially generated by zero padding~\cite{alsallakh2020mind}. When a convolutional kernel extracts the feature at the boundary with zero padding, the sharp transitions between the zero values and the original content are wrongly identified as an edge. Similarly, this edge effect can be induced by the deformed boundary of rectified images. As the convolutional layer increases, such an effect is gradually enlarged and constructs new features on the feature maps.

\begin{figure}[b]
\centering
\vspace{-0.5cm}
\includegraphics[width=1\linewidth]{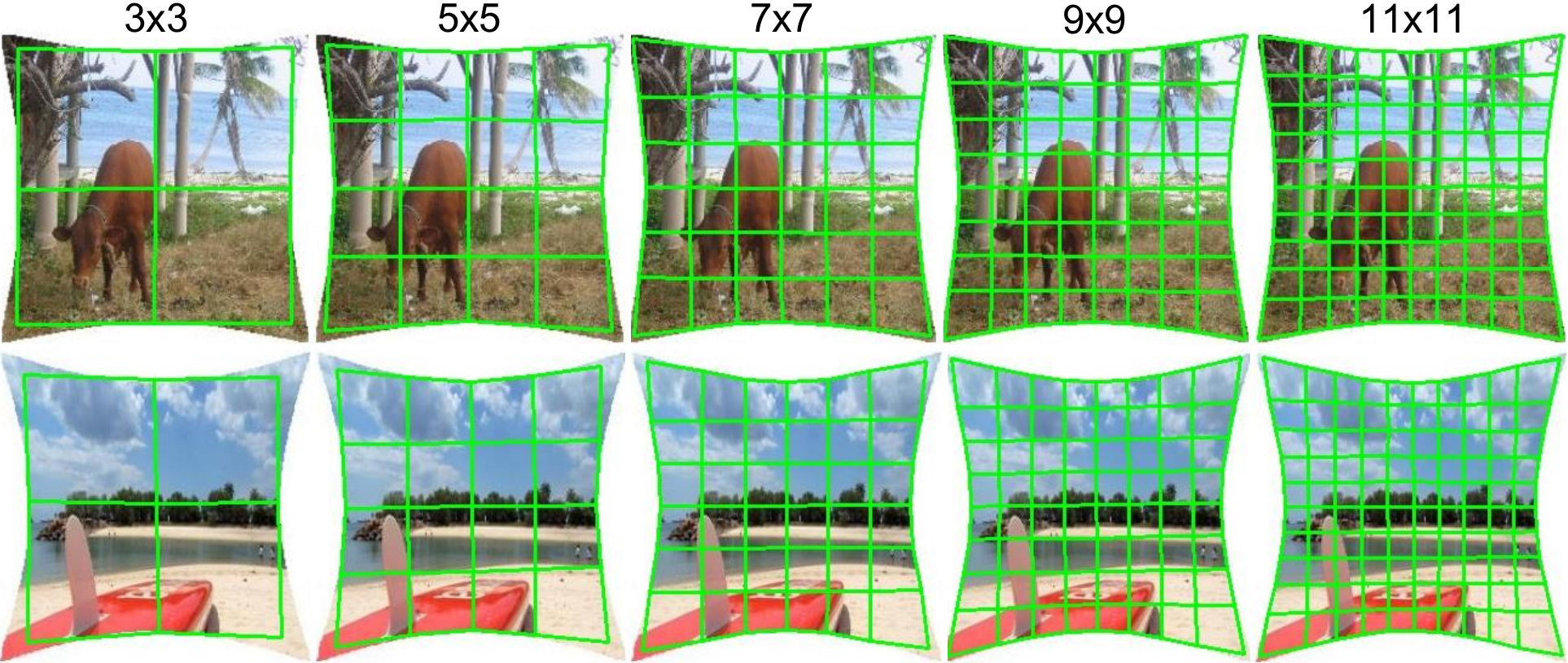}
\caption{Ablation study on the size of the control points in TPS.}
\vspace{-0.3cm}
\label{fig:ablation_grid}
\end{figure}

For the wrong perception case, the network mistakenly recognizes the boundary and its surrounding region as a surfboard. Since the edge effect introduces new features on the shallow feature maps and new semantics is formed in the deep feature maps. More results are reported in the supplementary material. Furthermore, the deformed boundaries not only exist in the rectified image, but also in other research regions such as image warping, image stitching, and roll shutter correction. Thus, the edge extension effect can introduce new features onto feature maps and further influence the downstream perception performance.

\begin{table}
\begin{center}
\caption{Ablation study on the size of the control points in TPS.}
\label{tb_ablation_grid}
\begin{tabular}{c|cccc}
\hline
Size & PSNR $\uparrow$ & SSIM $\uparrow$ & FID $\downarrow$ & LPIPS $\downarrow$\\
\hline
\hline
$3\times3$ & 17.96 & 0.5134 & 25.00 &0.1347\\
$5\times5$ & 18.41 & 0.5288 & 21.49 &0.1221\\
$7\times7$ & 18.59 & 0.5385 & 19.62 &0.1153\\
$9\times9$ & \textbf{18.68} & \textbf{0.5450} & \textbf{19.01} &\textbf{0.1136} \\
$11\times11$ & 18.64 & 0.5392 & 19.16 &0.1139\\
\hline
\end{tabular}
\end{center}
\vspace{-0.3cm}
\end{table}

\subsection{Ablation Study}
\label{sec_ablation}
\noindent\textbf{Control Points:}
As shown in Table~\ref{tb_ablation_grid} and Figure~\ref{fig:ablation_grid}, the experiments demonstrate more control points can facilitate the structure approximation and localization ability of RecRecNet for the deformed boundaries. And the size of $9\times9$ achieves the best performance on all evaluation metrics.

\begin{figure}[t]
\centering
\includegraphics[width=1\linewidth]{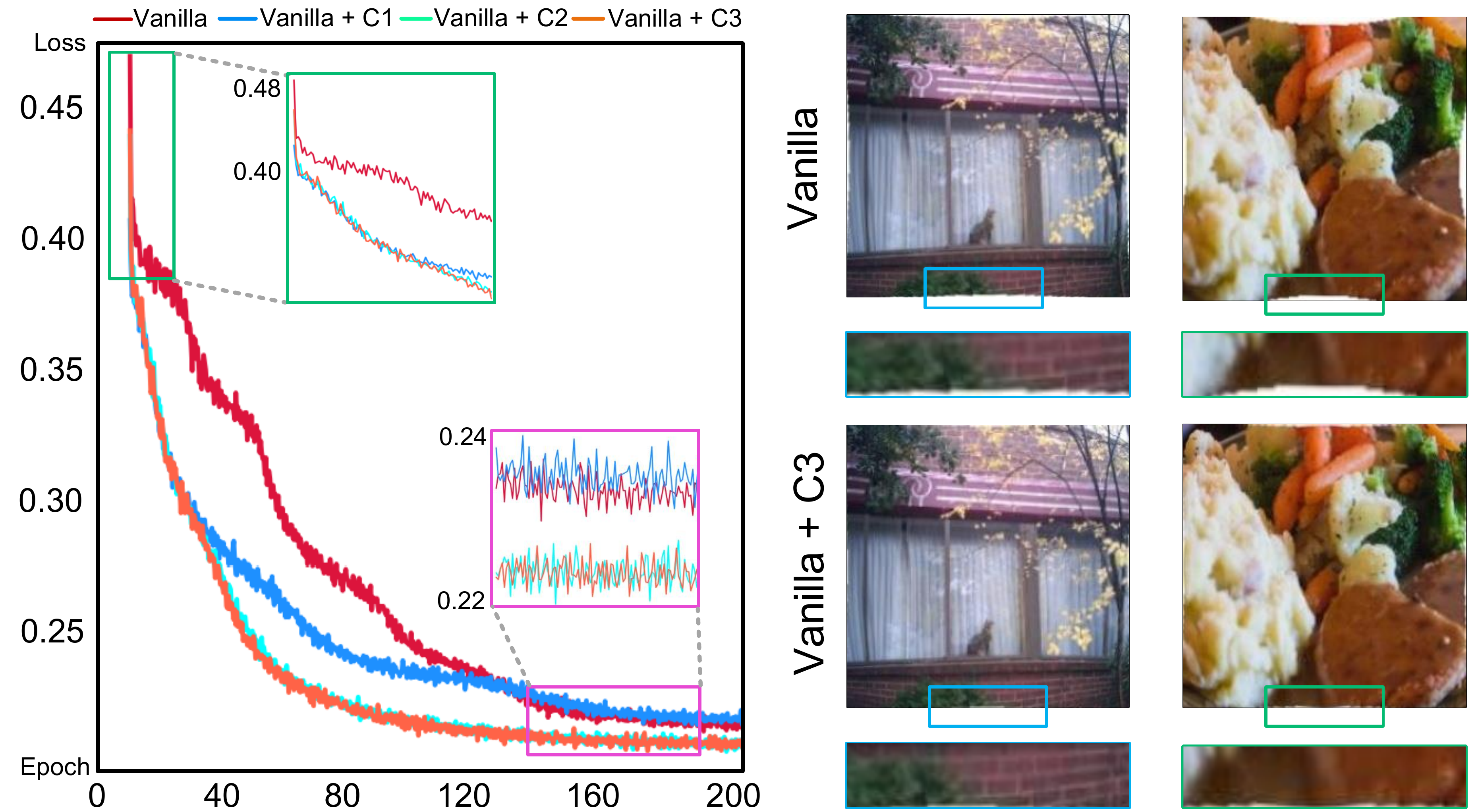}
\caption{Ablation study on the curriculum learning for rectangling task. C1, C2, and C3 denote different curriculums.}
\vspace{-0.2cm}
\label{fig:ablation_curriculum}
\end{figure}

\noindent\textbf{Curriculum Learning:} We define RecRecNet without a curriculum as Vanilla and define RecRecNet with different curriculums: 2-DoF $\mapsto$ 4-DoF as Vanilla + C1, 2-DoF $\mapsto$ 8-DoF as Vanilla + C2, and 4-DoF $\mapsto$ 8-DoF as Vanilla + C3, respectively. In Figure~\ref{fig:ablation_curriculum} (left), the training loss curves show that curriculum learning can facilitate a better initialization point for the training and faster convergence on the rectangling task. Vanilla training took around 40 more epochs to converge to a similar training performance as curriculum settings, but it is prone to overfitting on validation data. Moreover, a reasonable design for the curriculum (such as Vanilla + C3) helps to improve the boundary-aware ability of RecRecNet, enhancing the structure recovery performance as shown in Figure~\ref{fig:ablation_curriculum} (right).

\begin{figure}[t]
\centering
\includegraphics[width=.98\linewidth]{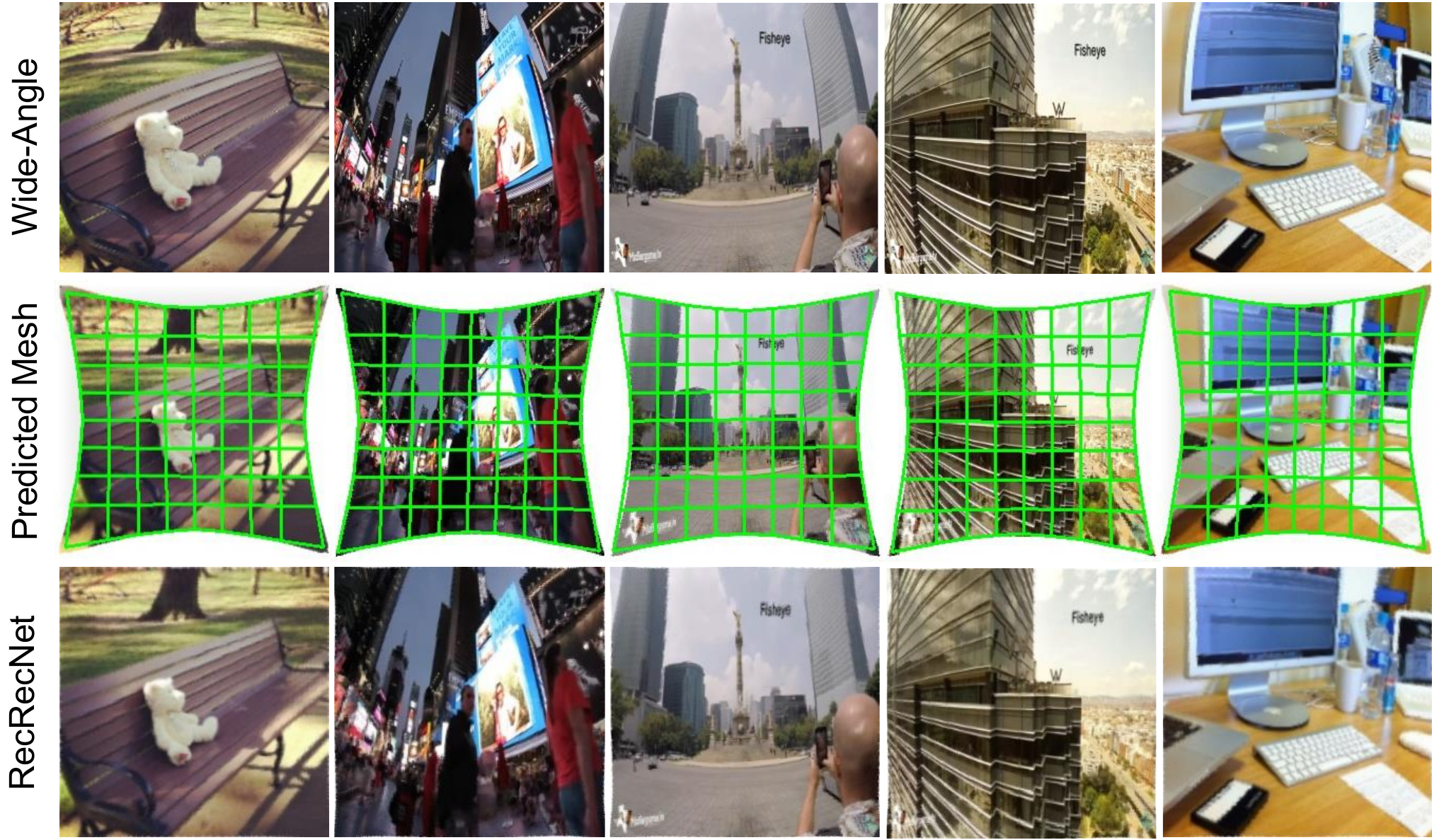}
\caption{Cross-domain evaluation on real-world wide-angle images and other types of synthesized datasets.}
\vspace{-0.2cm}
\label{fig:cross_domain}
\end{figure}

\begin{figure}[t]
\centering
\includegraphics[width=1\linewidth]{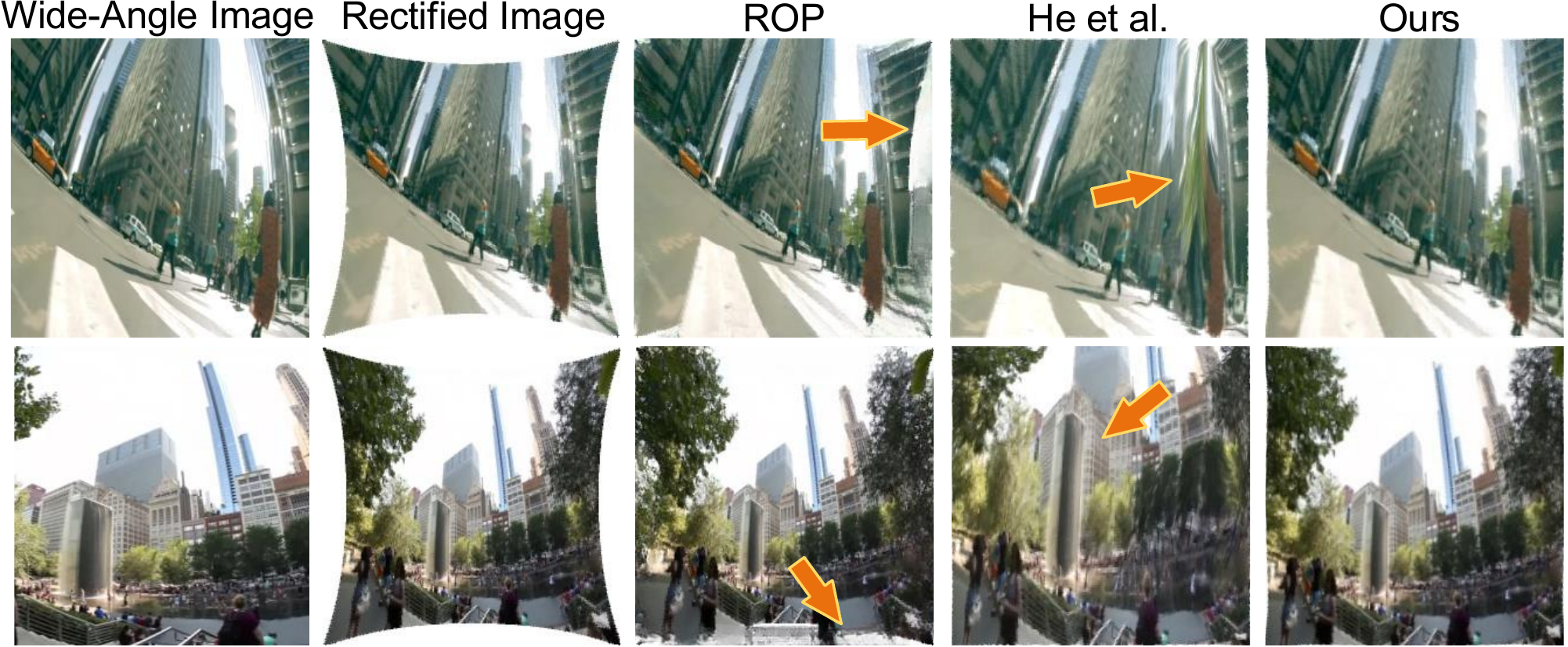}
\caption{Comparison results on the cross-domain evaluation.}
\vspace{-0.3cm}
\label{fig:cross_domain_cp}
\end{figure}

\subsection{Cross-domain Evaluation}
To evaluate the generalization, we collect 300 rectified wide-angle image results from the state-of-the-art rectification methods~\cite{OrdianlDistortion, PCN}. Their results are derived from various types of datasets and real-world wide-angle lenses such as the Rokinon 8mm Cine Lens, Opteka 6.5mm Lens, and GoPro. These lenses are usually used to obtain the panoramic image. Figure~\ref{fig:cross_domain} shows RecRecNet can well generalize to other domains and different rectified structures while it trained using only a synthesized image dataset with one type of camera model. We also show some comparison results as shown in Figure~\ref{fig:cross_domain_cp}, where fractured semantics and distorted object can be observed in previous methods.

\section{Conclusion}
\label{sec:conclusion}
In this work, we consider a new rectangling rectification task for the wide-angle image. While it has never been studied in previous literature, we demonstrate that the deformed boundary of the prevalent rectified wide-angle image can significantly influence the vision perception models. To eliminate this issue, we present to construct a win-win representation for the rectified wide-angle image and design a novel RecRecNet. Equipped with a flexible TPS transformation motion model, RecRecNet can formulate the local deformation from the deformed boundary to the straight one in an unsupervised end-to-end manner. Besides, we inspire RecRecNet to learn the gradual deformation rules with a DoF-based curriculum learning, which can relieve the complexity of non-linear and non-rigid transformation. Furthermore, a detailed analysis is provided to explain why the deformed image boundary makes the current vision perception deformed. In future work, we plan to extend to a general paradigm for rectangling any deformed images and further study the relationship between the image boundary and vision perception performance. Moreover, embedding our rectangling algorithm into an online data augmentation for training vision models would be also interesting.

{\small
\normalem
\bibliographystyle{ieee_fullname}
\bibliography{egbib}
}
\clearpage
\appendix

\section{Supplemental Material}
\subsection{Overview}
\label{s1}
In this document, we provide the following supplementary content:

\begin{itemize}
\item Details of justification for rectangling (Section \ref{s-j}).
\item Details of the dataset construction (Section \ref{s2}).
\item More qualitative results of the comparison methods and our approach (Section \ref{s3}).
\item More vision perception results (Section \ref{s4}).
\item More cross-domain evaluation results (Section \ref{s5}).
\item User study (Section \ref{s6}).
\end{itemize}

\subsection{Justification for Rectangling}
\label{s-j}
In the main manuscript, we have demonstrated that our rectangling method can significantly facilitate the downstream vision tasks, improving the performance of object detection and semantic segmentation. Besides the benefits of vision tasks, we argue there are other reasons why the rectangling algorithm is worth investigating. First, rectangling allows a visually pleasant structure for humans and a normal format for devices. Previous literature~\cite{he2013rectangling, he2013content} revealed most users prefer rectangular boundaries for publishing, sharing, and printing photos. For example, over 99\% images in the tag “panorama” on Flickr (flickr.com) have rectangular boundaries. And human vision system is more sensitive to irregular lines. Moreover, the rectangular image well fits the mainstream display window and screen. Instead, the blank region in a nonrectangular image occupies invalid space, which makes the data storage/compression inefficient.

\begin{figure*}[t]
\centering
\includegraphics[width=1\linewidth]{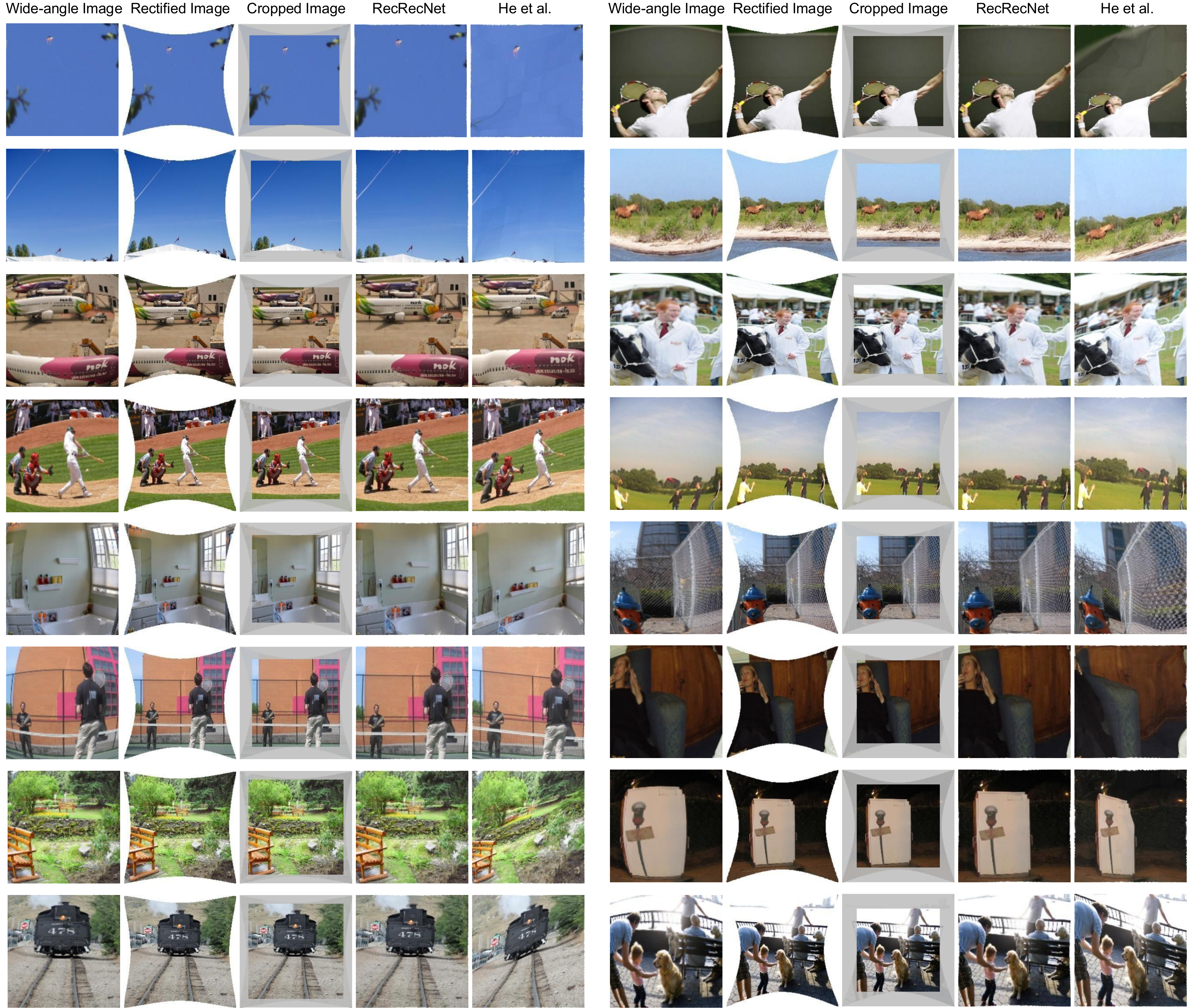}
\caption{More qualitative results compared to He et al.~\cite{he2013rectangling}. We show the wide-angle image, rectified image, cropped rectified image, and the rectangling result of our RecRecNet, as well as the rectangling result by He et al.~\cite{he2013rectangling} from left to right.}
\label{fig:more_res}
\end{figure*}

\subsection{Details of the Dataset Construction}
\label{s2}
We construct a rectangling rectification dataset that severing the first dataset in the research region, and we would like to release it to promote the research development. In particular, our dataset is built using the following four steps: \textbf{(i) Wide-angle image and rectified image synthesis}. Since it is extremely challenging to collect large-scale paired wide-angle images and their rectified ground truth, we follow the existing distortion rectification approaches~\cite{Rong, DeepCalib, FishEyeRecNet, DR-GAN, PCN} to synthesize the dataset. First, the original images are collected from the MS-COCO dataset~\cite{lin2014microsoft}. We leverage a $4^{th}$ order polynomial model to approximate the radial distortion of the wide-angle image, which is verified to meet most projection models with high accuracy. To be specific, four distortion parameters are randomly generated from the following ranges: $k_1 \in [-1\times10^{-4}, -1\times10^{-8}]$, $k_2 \in [1\times10^{-12}, 1\times10^{-8}]$ or $\in [-1\times10^{-8}, -1\times10^{-12}]$, $k_3 \in [1\times10^{-16}, 1\times10^{-12}]$ or $\in [-1\times10^{-12}, -1\times10^{-16}]$, and $k_4 \in [1\times10^{-20}, 1\times10^{-16}]$ or $\in [-1\times10^{-16}, -1\times10^{-20}]$. Then we perform the distortion rectification on the wide-angle image and obtain the rectified images. \textbf{(ii) Rectangling the rectified image}. Formulating an accurate transformation model for rectangling the rectified image is difficult due to its non-linear and non-rigid characteristics. We notice that there is a classical panoramic image rectangling technique He et al.~\cite{he2013rectangling} on computer graphics, it makes the stitched image regular by optimizing an energy function with line-preserving mesh deformation. Thus, we perform the same energy function on our rectified image dataset to fit the rectangling rectification task. However, the capability to preserve linear structures in He et al.~\cite{he2013rectangling} is limited by line detection. Consequently, some rectangling rectified images have nonnegligible distortions. To overcome this issue, we carefully filter all rectangling results and repeat the selection process three times, resulting in 5,160 training data from 30,000 source images and 500 test data from 2,000 source images. Each manual operation takes around 10s. The size of all images is $256\times256$. \textbf{(iii) Cross-domain validation}. In addition to the synthesized dataset, we collect 300 rectified wide-angle image results from the state-of-the-art rectification methods~\cite{liao2021multi, PCN}. Their results are derived from other types of datasets and real-world wide-angle lenses such as the Rokinon 8mm Cine Lens, Opteka 6.5mm Lens, and GoPro. \textbf{(iv) DoF-based curriculum dataset}. To relieve the challenge of the structure approximation in the rectangling task, we proposed a Degree of Freedom (DoF)-based curriculum learning. Specifically, three curriculum stages are leveraged to inspire our RecRecNet, namely, from similarity transformation (4-DoF) to homography transformation (8-DoF), and to rectangling transformation. Thus, we also construct two datasets (4-DoF dataset and 8-DoF dataset) to provide the basic transformation knowledge, in which each dataset contains 5,000 image pairs. For the transformation synthesis, we randomly perturb four corners of the original image and warp it to the target image following the previous work~\cite{detone2016deep}.

\subsection{More Qualitative Comparison Results}
\label{s3}
As shown in Figure~\ref{fig:more_res}, we exhibit more qualitative comparison results. Our RecRecNet is capable of rectangling the rectified wide-angle image in various scenes. We can observe the deformed boundary is straightened and the image content is rearranged to keep undistorted by RecRecNet, contributing a win-win rectification representation for the wide-angle image. By contrast, previous methods fail to trade off the image boundary and image content in the rectified image. Their results usually show incomplete content or distorted distributions. For example, some rectangling results produced by He et al.~\cite{he2013rectangling} rotate the original scene and twist the object due to their line-preserving mesh deformation.

\begin{figure*}[t]
\centering
\includegraphics[width=1\linewidth]{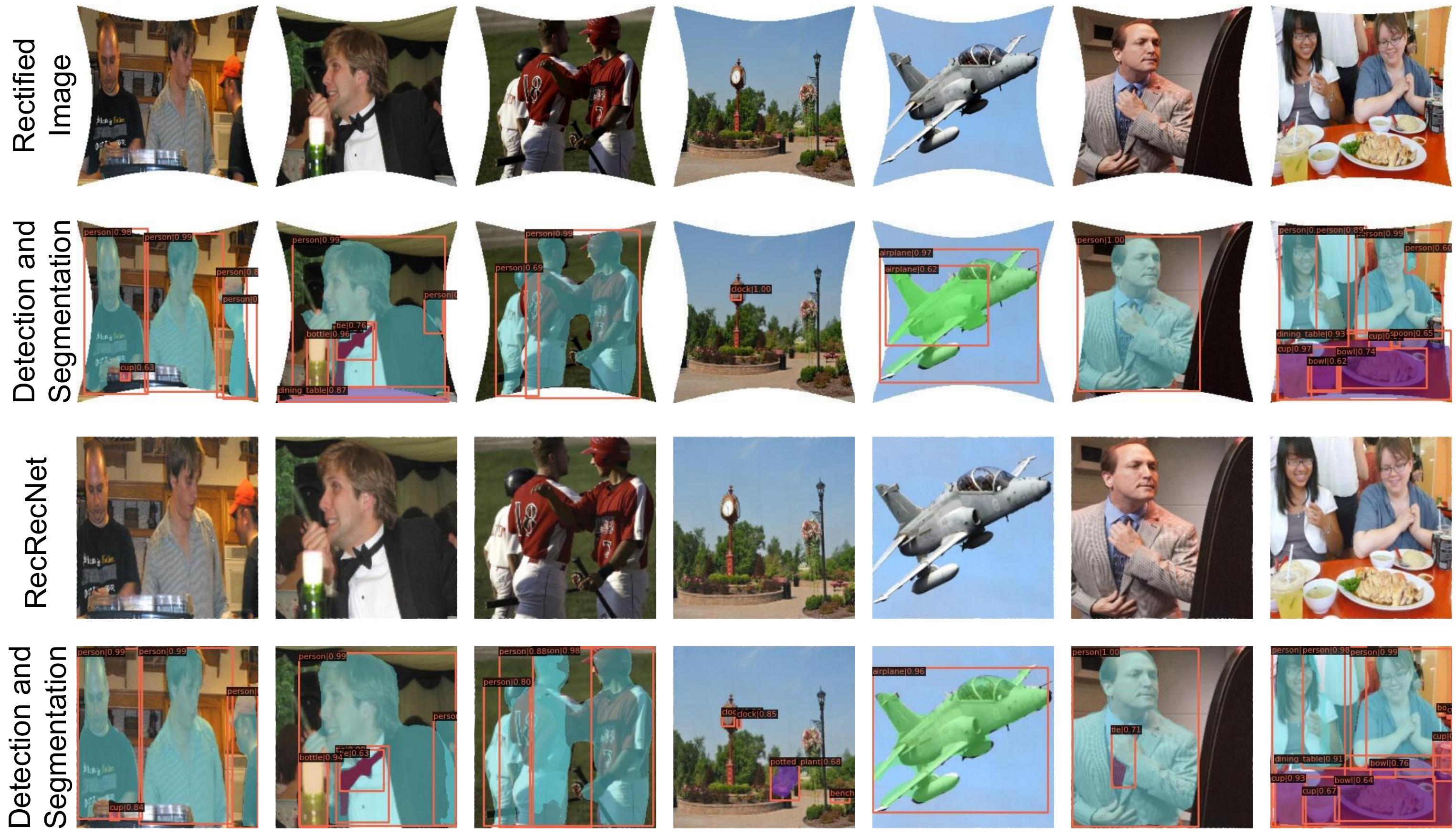}
\caption{More detection and segmentation results by Mask R-CNN~\cite{he2017mask}. We show the rectified wide-angle image and its vision perception result, and our rectangling result and its vision perception result, from top to bottom.}
\label{fig:more_perception}
\end{figure*}

\begin{figure}[t]
\centering
\includegraphics[width=1\linewidth]{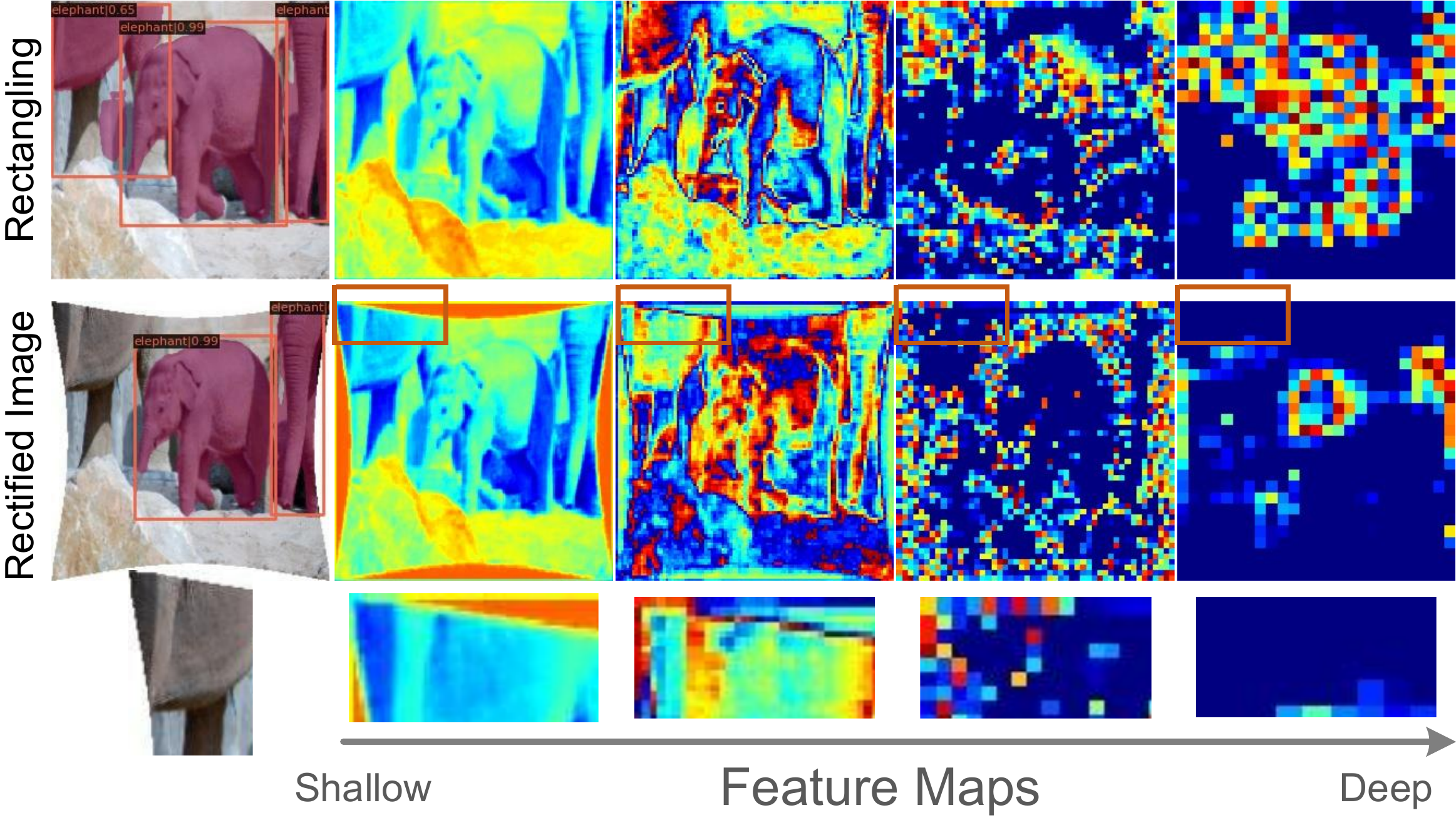}
\caption{Failure case of Mask R-CNN~\cite{he2017mask} in regards to the missing perception. We find that the deformed boundary can introduce new features to the original feature maps. As a result, the network cannot recognize the elephant near the deformed image boundary as the original features are blinded in the deep feature maps.}
\label{fig:missing_perception}
\end{figure}

\begin{figure*}[t]
\centering
\includegraphics[width=.95\linewidth]{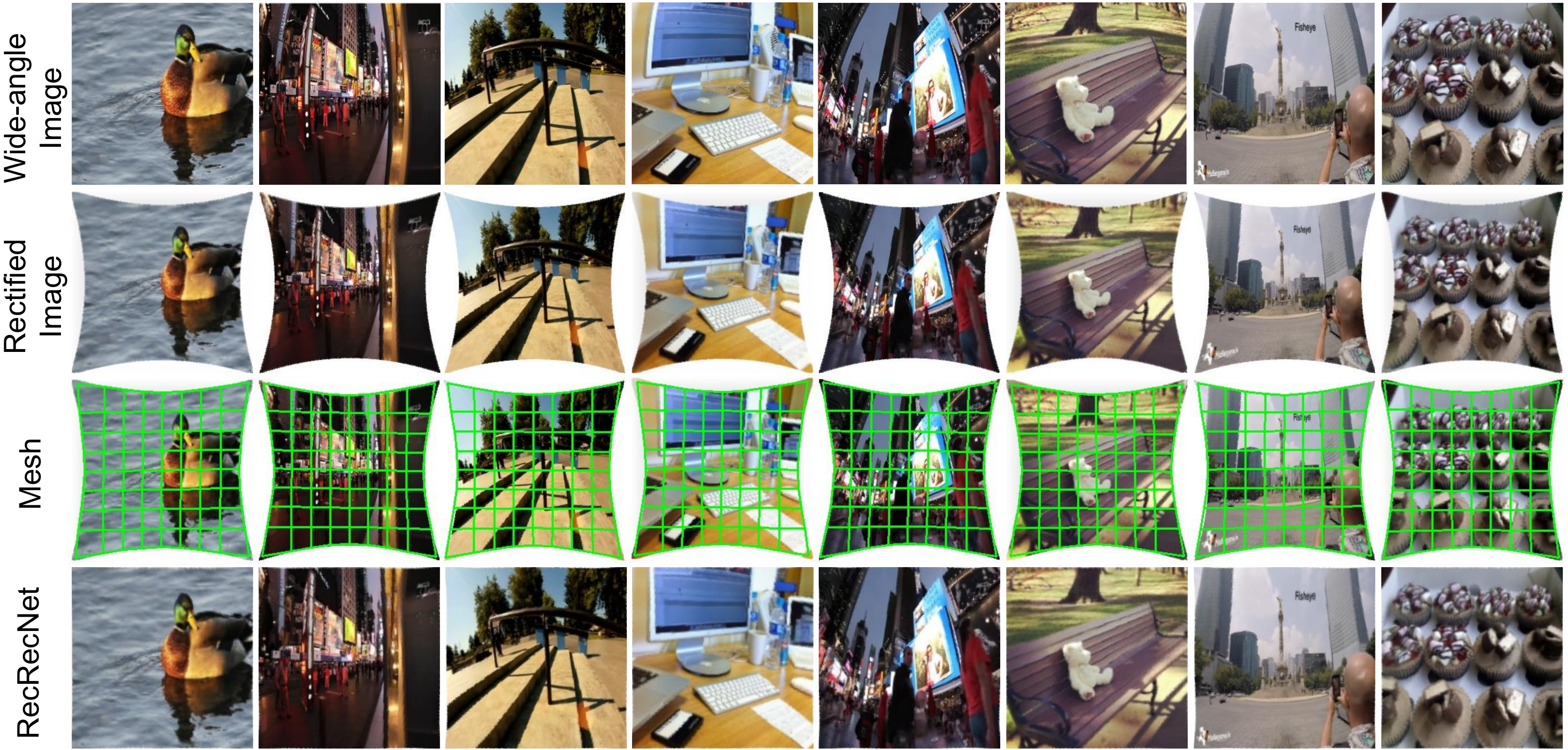}
\caption{More results for the cross-domain evaluation. We show the wide-angle image, rectified image, predicted mesh, and rectangling result of our RecRecNet from top to bottom.}
\label{fig:more_cross}
\end{figure*}

\subsection{More Vision Perception Results}
\label{s4}
As illustrated in Figure~\ref{fig:more_perception}, we show more vision perception results, including the object detection and semantic segmentation results, which are derived by the popular perception model Mask R-CNN~\cite{he2017mask}. As we can observe, the learning model is misled by the deformed structure in the original rectified image, leading to wrong perception and missing perception results. As a benefit of the proposed flexible TPS transformation module and DoF-based curriculum learning, our RecRecNet can significantly help the downstream vision tasks by eliminating the deformed boundary issue. And the perception performance is recovered especially near the image boundary. For the performance degradation in rectified images, one straightforward reason is that the perception model has never seen the data with a deformed image boundary during training, and the domain gap between test data and training data confuses it on rectified images. Nevertheless, we would like to explore deeper reasons and disclose why the deformed geometry makes the vision perception deformed.

\begin{figure}[t]
\centering
\includegraphics[width=1\linewidth]{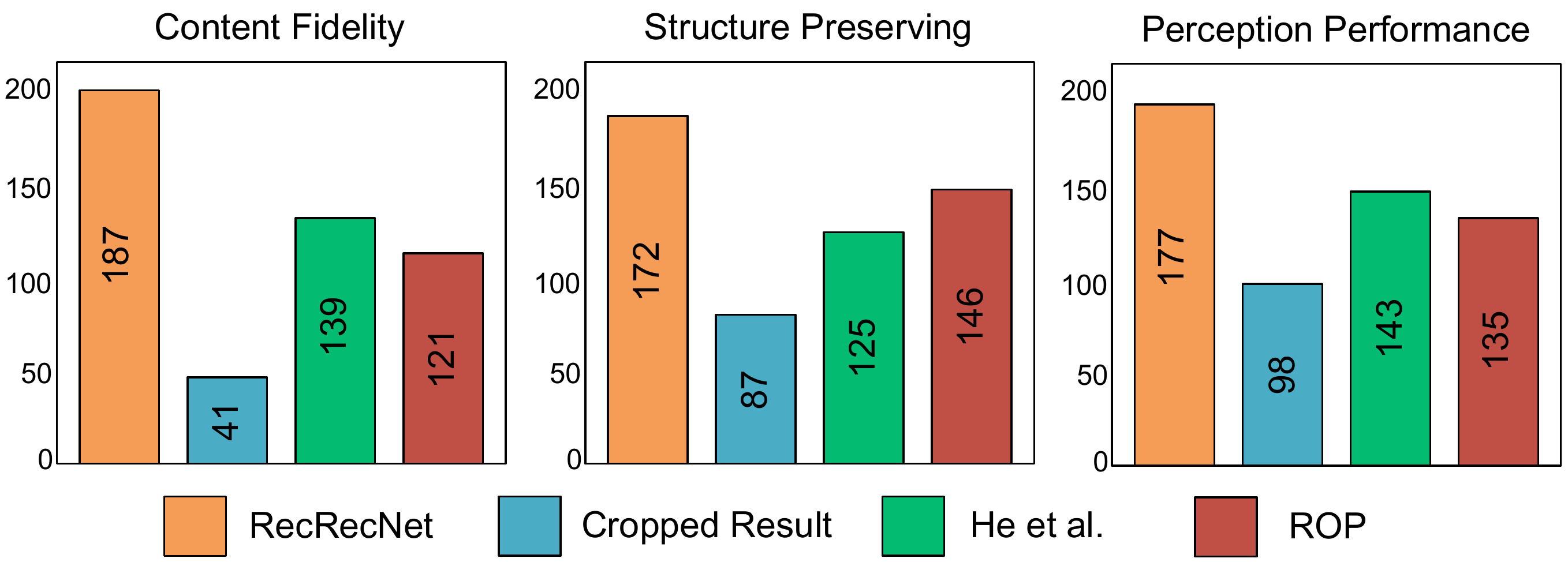}
\caption{User study for the rectangling rectification results.}
\label{fig:user_study}
\end{figure}

A sample is shown in Figure~\ref{fig:missing_perception} to visualize why the deformed image boundary makes the perception deformed. As described in the main manuscript, we found the deformed boundary can introduce \textit{new features} onto the feature maps, which (i) form new semantics (leads to the wrong perception) or (ii) cause blind spots for original features (leads to missing perception). The noticeable line artifacts and curve artifacts can be observed at the outermost boundaries and deformed boundaries respectively in shallow feature maps. In particular, the line artifacts are essentially generated by zero padding. Zero padding is a fundamental component of prominent CNNs architectures, it serves to maintain the size of the feature maps across the network by adding zero values to the feature map. When a convolutional kernel extracts the feature at the boundary with zero padding, the sharp transitions between the zero values and the original content are wrongly identified as an edge. Compared to the regular image, the rectified image allows a faster extension of the edge effect. Thus, more new features will be involved in the inference process. We call it a dual-edge extension effect.

Recent works also demonstrated that the zero padding can unintentionally leak the positional information in CNNs \cite{islam2019much,kayhan2020translation,xu2021positional}. In the case of the rectified image with a deformed boundary, we reason that the original positional information could be disturbed by the introduced new features. As a result, the perception model cannot build an accurate spatial distribution and fails to understand the relationships among adjacent semantics. Furthermore, it is interesting to find that the new semantic features introduced by the deformed boundary are jointly understood with their near features. In other words, the neural networks perceive the object relation and the scene context. 

For the missing perception case, if the features near the boundary are non-salient, the newly introduced features can cover them and generate blind spots for the perception model. As a result, the network cannot recognize the elephant near the deformed image boundary. We argue that such an effect can raise in most image transformation cases such as image warping, in which the boundaries are deformed to trade off different requirements on image content.

\subsection{More Cross-Domain Evaluation Results}
\label{s5}
As mentioned in Section~\ref{s2}, we collect 300 rectified wide-angle image results from the state-of-the-art rectification methods~\cite{liao2021multi, PCN} to conduct the cross-domain evaluation. Their results are derived from other types of synthesized datasets and real-world wide-angle lenses with different camera models. Figure~\ref{fig:more_cross} shows the cross-domain evaluation results. While the new data domain has never been seen, our RecRecNet can achieve a promising generalization ability by learning a flexible TPS transformation. And we can observe that the rectangling results display reasonable global distributions and visually pleasing local details. Moreover, the predicted mesh locates at the most spatial range of the rectified image, demonstrating the effectiveness of the learned rectangling transformation.

\subsection{User Study}
\label{s6}
The aim to yield the rectangular images is mainly for the visual sense and vision perception, thus we conduct a user study to evaluate the rectangling methods based on three aspects: content fidelity, structure-preserving, and perception performance. In particular, content fidelity denotes the measurement of the image content, such as the objects and background. Structure-preserving requires evaluation, especially for the rectangling image boundary, namely, if it is straight or not. Perception performance means the intuitive object detection and semantic segmentation results. Subsequently, we collect 200 samples in random order, and 10 volunteers with vision expertise are required to vote on the results under different contexts. As shown in Figure~\ref{fig:user_study}, RecRecNet achieves the highest votes in three tests and shows a superior capacity for scene faithfulness.

\end{document}